\documentclass[sigconf]{acmart}
\renewcommand\footnotetextcopyrightpermission[1]{}
\AtBeginDocument{%
  }

\setcopyright{acmlicensed}
\copyrightyear{2018}
\acmYear{2018}
\acmDOI{XXXXXXX.XXXXXXX}
\acmConference[Conference acronym 'XX]{Make sure to enter the correct
  conference title from your rights confirmation email}{June 03--05,
  2018}{Woodstock, NY}
\acmISBN{978-1-4503-XXXX-X/2018/06}

\usepackage{multirow}
\usepackage{subcaption}
\usepackage{booktabs}
\usepackage{makecell}
\usepackage{threeparttable}
\usepackage{enumitem}
\usepackage[table]{xcolor}
\usepackage{colortbl}
\usepackage{array}
\begin{document}

\title{MuDD: A Multimodal Deception Detection Dataset and GSR-Guided Progressive Distillation for Non-Contact Deception Detection}

\author{Peiyuan Jiang}
\affiliation{%
  \institution{School of Computer Science and Engineering, University of Electronic Science and Technology of China}
  \city{Chengdu}
  \state{Sichuan}
  \country{China}}
\email{darcy981020@gmail.com}

\author{Yao Liu}
\authornotemark[1]
\affiliation{%
  \institution{School of Information and Software Engineering, University of Electronic Science and Technology of China}
  \city{Chengdu}
  \state{Sichuan}
  \country{China}}
\email{liuyao@uestc.edu.cn}

\author{Yanglei Gan}
\affiliation{%
  \institution{School of Computer Science and Engineering, University of Electronic Science and Technology of China}
  \city{Chengdu}
  \state{Sichuan}
  \country{China}}
\email{yangleigan@std.uestc.edu.cn}

\author{Jiaye Yang}

\affiliation{%
  \institution{School of Computer Science and Engineering, University of Electronic Science and Technology of China}
  \city{Chengdu}
  \state{Sichuan}
  \country{China}}
\email{202411081710@std.uestc.edu.cn}

\author{Lu Liu}

\affiliation{%
  \institution{School of Computer Science and Engineering, University of Electronic Science and Technology of China}
  \city{Chengdu}
  \state{Sichuan}
  \country{China}}
\email{202522080827@std.uestc.edu.cn}

\author{Daibing Yao}
\affiliation{%
  \institution{Yizhou Prison, Sichuan Province}
  \city{Chengdu}
  \state{Sichuan}
  \country{China}}
\email{357497551@qq.com}

\author{Qiao Liu}

\affiliation{%
  \institution{School of Computer Science and Engineering, University of Electronic Science and Technology of China}
  \city{Chengdu}
  \state{Sichuan}
  \country{China}}
\email{qliu@uestc.edu.cn}

\thanks{*Corresponding author: Yao Liu}

\renewcommand{\shortauthors}{Jiang et al.}
\begin{abstract}
Non-contact deception detection remains challenging because visual and auditory deception cues often lack stable cross-subject patterns. To improve non-contact deception detection, we explore cross-modal knowledge distillation (CMKD) from galvanic skin response (GSR), which provides more reliable physiological evidence of deception, to guide representation learning in non-contact modalities. To enable this study, we introduce MuDD, a large-scale multimodal deception detection dataset collected under the Guilty Knowledge Test (GKT) paradigm, with recordings from 130 participants spanning 690 minutes across six modalities. 
Building on MuDD, we further propose GSR-guided Progressive Distillation (GPD), a cross-modal distillation framework that models cross-modal knowledge transfer as a dynamic process. Unlike existing CMKD methods that typically rely on predefined distillation schemes, GPD adaptively regulates both the transfer pathway and the transferred knowledge according to the evolving representational gap between teacher and student. Specifically, GPD uses gap-aware dynamic routing to select suitable distillation configurations based on the evolving representational gap between teacher and student, and progressively adjusts the relative importance of feature-level and logit-level knowledge during training to facilitate stable transfer across heterogeneous modalities. Extensive experiments and visualizations show that GPD outperforms existing methods and achieves state-of-the-art performance on both deception detection and concealed-digit identification. 

\end{abstract}




% \received{20 February 2026}
% \received[revised]{12 March 2026}
% \received[accepted]{5 June 2026}

\maketitle

\begin{figure}[ht]
    \centering
    %\begin{minipage}{0.48\textwidth}
        \centering
        \includegraphics[width=0.48\textwidth]{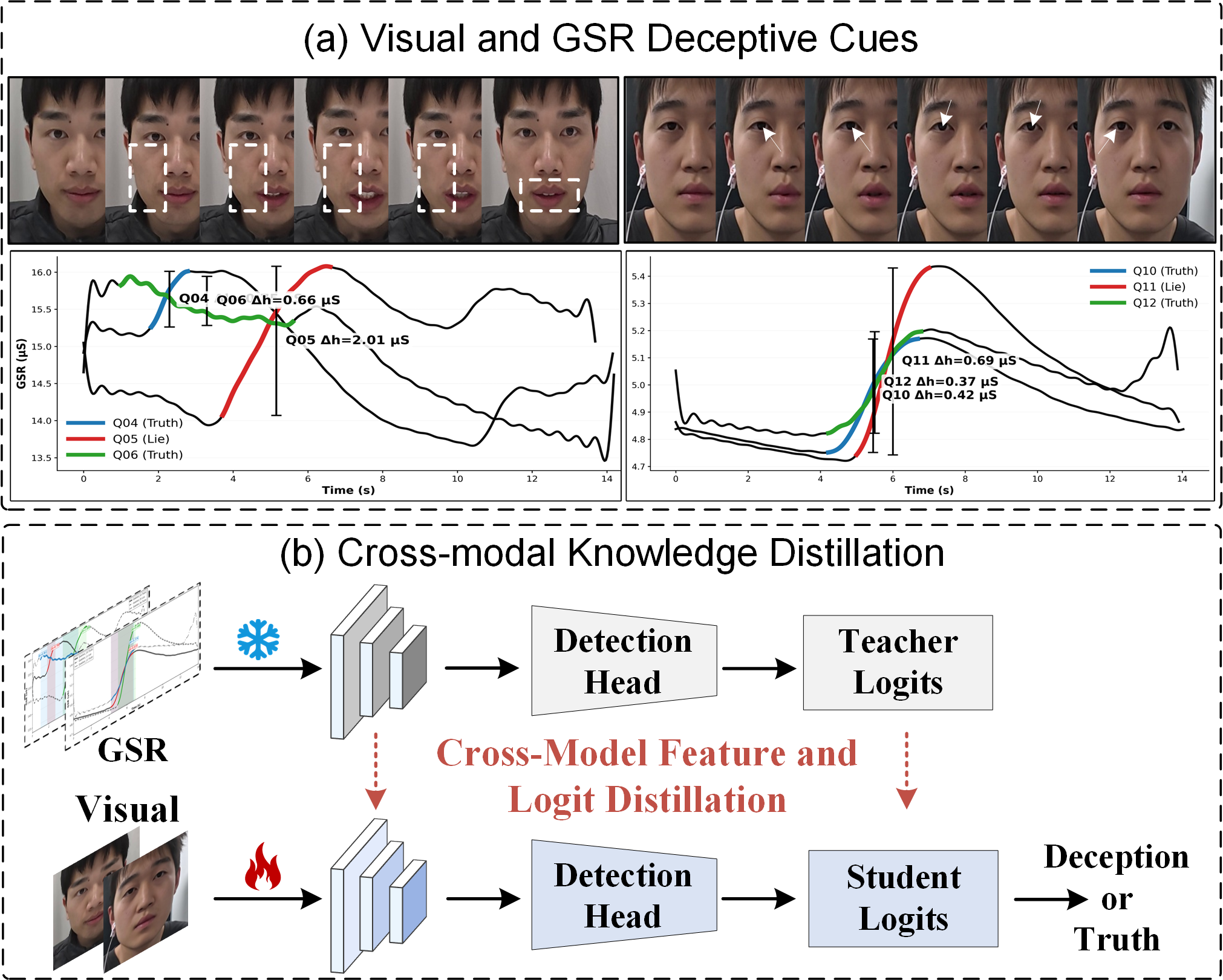}  % 调整图片宽度为页面宽度的45%
    \caption{Motivation of this work. (a) Visual deception cues are often subtle and subject-dependent, making it difficult to identify stable cross-subject patterns, whereas GSR exhibits more consistent deception-related responses across individuals. (b) Illustration of the cross-modal knowledge distillation paradigm, where GSR serves as the teacher modality to transfer feature-level and logit-level knowledge to non-contact modalities through a teacher--student framework.} 
    \label{fig:fig1}  % 标签用于引用
    % \end{minipage}
\end{figure}

\begin{table*}[t]
\centering
\caption{Comparison of MuDD with existing deception-related datasets. Abbreviations: Subs. = subjects, Dur. = duration, V = video, A = audio, T = text, Pers. = personality traits, Bin. = binary label, and T-10cls = trial-level 10-class label.}
\label{tab:dataset_comparison}
\setlength{\tabcolsep}{7pt}
\renewcommand{\arraystretch}{1}
\resizebox{\textwidth}{!}{
\begin{tabular}{lccccc}
\toprule
\rowcolor{gray!15}
\textbf{Dataset} & \textbf{Subs.} & \textbf{Dur.} & \textbf{Setting} & \textbf{Modalities} & \textbf{Labels} \\
\midrule

\rowcolor{gray!8}
\multicolumn{6}{l}{\textbf{\textit{Without physiological signals}}} \\
CSC Deceptive Speech~\cite{ref23} & 32 & 7920 min & Interview & A & Bin. \\
CXD Corpus~\cite{ref24} & 340 & 7200 min & False interview & A+T & Bin. \\
Real-life Trial~\cite{ref25} & 56 & $< 60$ min & Courtroom & V+T & Bin. \\
MU3D~\cite{ref27} & 80 & $< 200$ min & Social interview & V+T & Bin. \\
Truth-Tellers~\cite{ref26} & 60 & $< 480$ min & Simulated interview & V & Bin. \\
Box of Lies~\cite{ref11} & 26 & 144 min & Bluffing game & V+T & Bin. \\
DOLOS~\cite{ref29} & 213 & $\approx 530$ min & Daily deception game & V+A & Bin. \\
MDPE~\cite{ref30} & 193 & 6209 min & Simulated interview & V+A+T+Pers. & Bin. \\

\midrule
\rowcolor{gray!8}
\multicolumn{6}{l}{\textbf{\textit{With physiological signals}}} \\
GSR Deception Dataset~\cite{ref53} & 20 & $< 320$ min & Bluffing game & GSR & Bin. \\
Bag of Lies~\cite{ref31} & 35 & $\approx 110$ min & Daily lying & V+A+EEG+Gaze & Bin. \\
SEUMLD~\cite{ref13} & 76 & $\approx 366$ min & Simulated interview & V+A+ECG & Bin. \\
CogniModal-D~\cite{ref32} & $\approx 110$ & -- & Scenario task & EEG+ECG+Gaze+GSR+A+T & Bin. \\

\rowcolor{gray!12}
\textbf{MuDD (ours)} & \textbf{130} & \textbf{690 min} & \textbf{GKT} & \textbf{V+A+GSR+PPG+HR+Pers.} & \textbf{Bin.+T-10cls} \\
\bottomrule
\end{tabular}
}
\end{table*}

\section{INTRODUCTION}

Deception is a fundamental yet elusive aspect of human social behavior, with important implications for security screening, forensic investigation, and interpersonal communication. Despite its practical significance, unaided human ability to detect deception remains only slightly above chance level \cite{ref1,ref2}, highlighting the inherent difficulty of relying on subjective judgment alone. This limitation has driven the development of more objective approaches for deception assessment. Early studies primarily relied on contact-based polygraph methods, which infer deceptive behavior by measuring physiological responses associated with autonomic nervous system activity, such as respiration, heart rate (HR), and galvanic skin response (GSR), under standardized testing protocols \cite{ref3,ref4}.

Although contact-based polygraph methods have demonstrated relatively high reliability, their practical applicability remains limited by the need for specialized equipment and controlled examination procedures. To address these limitations, recent research has increasingly turned to non-contact deception detection, which seeks to infer deceptive behavior from overt and non-invasive signals, such as facial expressions and linguistic cues \cite{ref7,ref8,ref9}. However, the performance of existing non-contact methods on public benchmarks remains modest and inconsistent across datasets \cite{ref11,ref30,ref13}, with reported performance typically ranging from 63\% to 73\%\footnote{Performance denotes either accuracy or recall.}. Although these results are not directly comparable due to differences in datasets and evaluation metrics, their substantial variation and overall moderate performance nevertheless reflect the difficulty of non-contact deception detection.

One important reason is that deception cues in audio-visual modalities are often subtle and highly subject-dependent, making it difficult to learn stable and generalizable patterns across individuals \cite{ref51}. As illustrated in Fig. \ref{fig:fig1}(a), two deceptive cases from different subjects exhibit markedly different visual characteristics. The subject on the left shows relatively pronounced facial changes under deception, whereas the subject on the right exhibits only weak visual variation, without clearly salient facial cues. This variability suggests that visual deception cues are often neither robust nor consistently shared across subjects. In contrast, the corresponding GSR signals exhibit a clearer cross-subject regularity, where the \textcolor{red}{\textbf{red}} curve denotes the GSR response to a deceptive question, while the \textcolor{green}{\textbf{green}} and \textcolor{blue}{\textbf{blue}} curves correspond to truthful responses to the immediately preceding and following questions. Although the absolute signal magnitude and waveform differ across subjects, the deceptive response consistently exhibits larger fluctuations than its adjacent truthful counterparts.

This observation raises a natural question: \textbf{\textit{can the more stable discriminative patterns in GSR be exploited to improve representation learning in non-contact modalities?}} A natural paradigm for addressing this question is cross-modal knowledge distillation (CMKD). As illustrated in Fig.~\ref{fig:fig1}(b), CMKD provides a general teacher--student framework in which GSR can serve as the teacher modality, while the target non-contact modality is treated as the student. Through distillation objectives, the student is encouraged to absorb discriminative knowledge from GSR, thereby potentially improving deception detection performance when only the non-contact modality is available at test time \cite{ref16}.

Although CMKD has demonstrated effectiveness in other domains~\cite{ref15,afouras2020asr,ref17,ref18}, its application to non-contact deception detection still faces two challenges. First, there is a lack of a large-scale dataset for studying how GSR can be leveraged to improve non-contact deception detection. Second, the large heterogeneity between GSR and non-contact modalities makes effective cross-modal transfer difficult. Existing CMKD methods have explored various ways to alleviate this heterogeneity, either by making teacher knowledge more transferable or by improving the transfer process~\cite{ref44,ref42,ref40,ref37,ref50,aslam2024distilling,ref46,ref40,ref41,ref18}. Despite their effectiveness, most of these methods do not explicitly model cross-modal distillation as a dynamically evolving process during training, which may limit effective knowledge transfer under large modality discrepancy.
To address these issues, we first construct a large-scale Multimodal Deception Detection Dataset (MuDD), containing 130 subjects and 690 minutes of recordings. In addition to the task-required video, audio, and GSR modalities, the dataset also provides Photoplethysmography (PPG), heart rate, and personality trait information, supporting a wide range of deception-related research. Based on MuDD, we further propose GSR-guided Progressive Distillation (GPD), a cross-modal distillation framework for non-contact deception detection. Unlike existing CMKD methods that typically rely on predefined distillation schemes, GPD models cross-modal distillation as a dynamic adaptation process. Specifically, it jointly incorporates two types of distillation knowledge and four candidate transfer configurations. According to the evolving representational gap between teacher and student, GPD uses gap-aware dynamic routing to adaptively select suitable transfer configurations, and employs progressive weighting to regulate the relative importance of feature-level and logit-level knowledge over training. Through this joint adaptation of transfer pathway and transferred knowledge, GPD facilitates cross-modal knowledge transfer.  

The main contributions of this work are summarized as follows:
\begin{itemize}
    \item We construct MuDD, a large-scale multimodal deception detection dataset collected under the classical Guilty Knowledge Test (GKT) paradigm, providing a benchmark for a wide range of automatic deception detection studies.
    \item We propose GPD, a GSR-guided progressive CMKD framework for non-contact deception detection. The proposed framework dynamically models the overall process of CMKD through gap-aware routing and progressive weighting.
    \item Extensive experiments and visualization analyses show that, compared with existing methods, GPD enables the student to learn teacher knowledge effectively, resulting in more teacher-aligned decision behavior and better performance.
\end{itemize}

\section{RELATED WORK}
\subsection{Multimodal Deception Detection Datasets}
Most existing deception detection datasets are designed for specific research questions. For example, \cite{ref23} introduced the Columbia SRI Colorado Deceptive Speech dataset to investigate the role of acoustic, prosodic, and lexical cues in deception recognition. Building on this, \cite{ref24} further incorporated textual information and cross-cultural factors to explore the potential of joint modeling of language and speech under cross-cultural conditions. \cite{ref25} introduced the Real-life Trial (RTL) dataset based on real courtroom videos to investigate visual and linguistic cues to deception in real-world settings. \cite{ref26,ref27,ref11,ref29,ref30} introduced datasets such as Truth-Tellers, MU3D, Box of Lies, DOLOS, and MDPE under diverse settings, including simulated interviews, social narratives, game-based interactions, and reward--penalty paradigms, providing important support for multimodal deception detection research. However, the absence of physiological signals limits their utility for studying the relationship between physiological arousal and overt behavioral cues during deception. Some datasets have introduced physiological signals such as electroencephalography (EEG) and electrocardiography (ECG) for cross-modal deception detection~\cite{ref31,ref13}, but they still lack galvanic skin response (GSR), a widely used and relatively reliable indicator in deception research~\cite{ref5}. The GSR Deception Dataset~\cite{ref53} was designed to examine whether GSR can capture deception- and suspicion-related states, while CogniModal-D~\cite{ref32} further extended this line of research by incorporating both GSR and overt behavioral signals. However, the former is limited to single-modality physiological data. The latter is not collected under classical polygraph-style paradigms such as the Control Question Test (CQT) or Guilty Knowledge Test (GKT), which are designed to induce deception-related autonomic arousal. As a result, these datasets remain limited for studying the relationship between GSR and non-contact modalities. In contrast, MuDD synchronously acquires GSR and non-contact modalities, including audio and video, under the classical GKT paradigm, thereby better supporting the study of GSR--overt modality relationships while also accommodating multimodal deception detection research. A comparison of MuDD with existing deception-related datasets is shown in Table~\ref{tab:dataset_comparison}.

\subsection{Cross-Modal Knowledge Distillation}
Knowledge distillation (KD) is an effective technique for transferring knowledge from one neural network to another~\cite{ref33}. Its core mechanism is a teacher--student framework, in which the student is trained to mimic the teacher by aligning teacher knowledge at different distillation locations, such as logits and intermediate features, under various objectives including divergence-based, regression-based, and relation-preserving losses~\cite{ref34,ref35,ref36,ref37}. Cross-modal knowledge distillation (CMKD) extends KD to the setting where the teacher and student receive inputs from different modalities~\cite{ref38}. A long-standing challenge in CMKD is that the large discrepancy between modalities often makes conventional KD methods less effective when directly applied to heterogeneous teacher--student pairs~\cite{ref40}. 
To alleviate modality heterogeneity, existing CMKD methods have explored a variety of strategies. Some methods improve transferability by reshaping teacher knowledge into forms that are less modality-specific and easier for the student to learn, such as knowledge decomposition~\cite{ref40,ref19}, shared-knowledge distillation~\cite{ref44,ref17}, intermediate feature alignment~\cite{ref15}, and relational or prototype-level distillation~\cite{ref50}. Other methods improve how knowledge is conveyed to the student by introducing better matched transfer mechanisms, such as bridging modules for easing cross-modal transfer~\cite{ref46}, proxy or assistant models~\cite{ref40,ref41}, and adaptive teacher--student assignment strategies, including adaptive teacher selection~\cite{ref42} and dynamic teacher--student matching~\cite{ref18}. 

While these methods have demonstrated promising effectiveness, most of them still rely on predefined distillation schemes and do not explicitly model CMKD as a dynamically evolving process throughout training. In practice, however, the teacher--student gap is not static. As training progresses, it may change substantially, which in turn affects both what knowledge is more learnable and what transfer process is more suitable at different stages. Motivated by this observation, we propose a dynamic CMKD framework that explicitly models cross-modal distillation as an evolving process during training. The framework adaptively regulates both transfer pathways and distillation signals according to the changing teacher--student representation gap. 
% The connections and differences between our method and existing CMKD methods are summarized in Fig.~\ref{fig:fig2}.
\begin{figure*}[t]
    \centering
    \includegraphics[width=\textwidth]{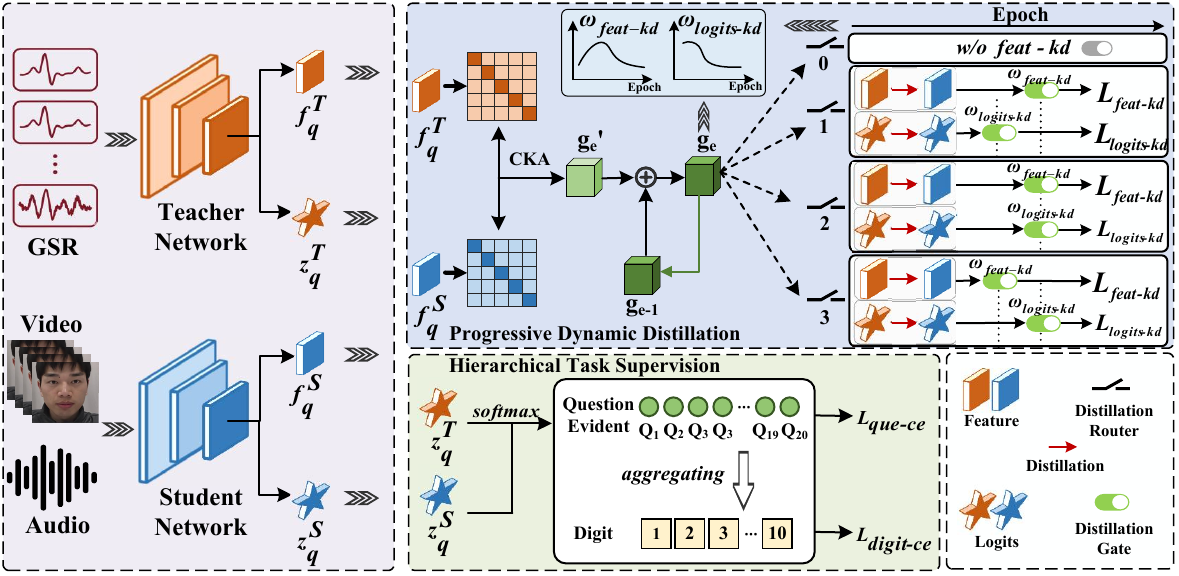}
    \caption{Overview of the proposed GSR-guided Progressive Distillation (GPD) framework.}
    \label{fig:gpd_framework}
\end{figure*}

\section{MuDD Dataset} 
\label{subsec:mudd}

The Multimodal Deception Detection (MuDD) dataset was collected under the Guilty Knowledge Test (GKT) paradigm, which assumes that concealed information elicits stronger physiological responses to probe stimuli than to irrelevant stimuli. Accordingly, we formulate the task as concealed-number identification. In each trial, a participant privately selects one target number from 1 to 10. During testing, the ten candidate numbers are presented in randomized order over two rounds, each containing one probe and nine irrelevant numbers. All questions are delivered through pre-recorded professional interrogation audio to standardize stimulus presentation and reduce unintended interference~\cite{ref47v1}, followed by a 10-second interval to capture the rise-and-recovery dynamics of the electrodermal response~\cite{lykken1959gsr,fowles1981publication}. Participants respond deceptively only to the selected number and truthfully to all others. Each participant contributes 20 dialogue clips, including 2 deceptive clips and 18 truthful clips, with both clip-level binary deception labels and trial-level 10-class concealed-number labels. The dataset includes synchronized visual, audio, and physiological recordings, together with Big-Five personality scores as subject-level trait annotations. Specifically, facial and upper-body behaviors are recorded at 30~fps, speech at 48~kHz, and physiological signals including GSR, PPG, and HR at 256~Hz. MuDD contains valid data from 130 native Chinese-speaking participants, totaling 690 minutes of synchronized multimodal recordings. All participants were recruited from a university student population. The study was approved by the institutional ethics review board, and written informed consent was obtained from all participants. The dataset will be publicly released for research purposes upon acceptance of this paper.

\section{Methodology}
\subsection{Problem formulation}
\label{subsec:problem}
For each subject, MuDD contains \(Q=20\) aligned question segments. We denote the student input sequence as \(\mathcal{X}=\{x_q\}_{q=1}^{Q}\), where each \(x_q\) denotes the feature representation of the target non-contact modality for segment \(q\), instantiated as either visual or audio features in our experiments, and the teacher GSR sequence as \(\mathcal{G}=\{g_q\}_{q=1}^{Q}\). Each segment is labeled with \(y_q\in\{0,1\}\), where \(1\) indicates denial of the concealed item and \(0\) indicates denial of an irrelevant item. At the trial level, each subject has a concealed digit label \(d^\star\in\{1,\dots,10\}\). Accordingly, our goal is to learn question-level deception detection from non-contact signals and, based on the aggregated deception evidence across the 20 questions, perform trial-level concealed-digit identification. 

\subsection{Overview of GPD}
An overview of the proposed GPD framework is shown in Fig.~\ref{fig:gpd_framework}. GPD consists of four components: a teacher network, a student network, a progressive knowledge distillation module, and a hierarchical task supervision module. The teacher takes GSR as input, while the student takes the target non-contact modality as input, instantiated as either visual or audio features in our experiments. For each question segment, both networks produce feature representations and deception logits for cross-modal knowledge transfer. The progressive distillation module defines four candidate routes for transferring feature-level and logit-level teacher knowledge. During training, dynamic routing selects the route according to the teacher--student gap, and progressive weighting adjusts the relative importance of the two knowledge types. The hierarchical task supervision module jointly performs question-level and concealed-digit-level supervision in one framework. Binary deception labels supervise question-level predictions, while the two question-level deception scores for each digit are aggregated into digit-level evidence for concealed-digit supervision.

\subsection{Teacher and student networks}
\label{subsec:arch}
The teacher and student networks share a similar prediction pipeline but operate on different modalities. For each segment \(q\), the teacher network \(\mathcal{T}\) takes the GSR input \(g_q\), while the student network \(\mathcal{S}\) takes the target non-contact input \(x_q\), instantiated as either visual or audio features in our experiments. They produce two-dimensional logit vectors and intermediate feature representations as
\begin{equation}
(z_q^T, f_q^T)=\mathcal{T}(g_q), \qquad
(z_q^S, f_q^S)=\mathcal{S}(x_q),
\end{equation}
where \(z_q^T,z_q^S\in\mathbb{R}^2\) are the output logits used for logit-level knowledge distillation, \(f_q^T\in\mathbb{R}^{D_T}\) and \(f_q^S\in\mathbb{R}^{D_S}\) are the intermediate feature embeddings used for feature-level knowledge transfer. The teacher and student deceptive-class probabilities are defined as
\begin{equation}
t_q=\mathrm{softmax}(z_q^T)_1, \qquad
p_q=\mathrm{softmax}(z_q^S)_1,
\end{equation}
where \(t_q\) and \(p_q\) denote the teacher and student probabilities of the deceptive class, respectively, and class label \(1\) denotes deception.

\subsection{Progressive knowledge distillation}
\label{subsec:progressive}
GPD distills two types of teacher knowledge: feature-level representations and logit-level predictions. Different scheduling patterns of these two knowledge types over the course of training define four route configurations, \(\mathcal{R}=\{0,1,2,3\}\), corresponding to no-feature, logit-first, joint, and feature-first distillation, respectively. Specifically, Route 0 disables feature-level distillation, Route 1 prioritizes logit-level distillation over feature-level distillation, Route 2 applies both jointly, and Route 3 prioritizes feature-level distillation over logit-level distillation. Dynamic routing selects the route configuration, and progressive weighting determines the relative contributions of the two knowledge types.
\paragraph{Gap-aware dynamic routing.}
Let \(F_e^T=\{f_q^T\}_{q=1}^{Q}\) and \(F_e^S=\{f_q^S\}_{q=1}^{Q}\) denote the teacher and student feature sets collected over the \(Q\) question segments at epoch \(e\), respectively. The routing gate at epoch \(e\) is the teacher--student gap \(g_e\), computed from the Centered Kernel Alignment (CKA) similarity between \(F_e^S\) and \(F_e^T\) as
\begin{equation}
\tilde g_e = 1-\mathrm{CKA}(F_e^S,F_e^T),
\end{equation}
\begin{equation}
g_e = \mu g_{e-1} + (1-\mu)\tilde g_e,
\end{equation}
where \(\tilde g_e\) denotes the raw teacher--student gap and \(\mu\in[0,1)\) is the EMA momentum. Given the gate signal \(g_e\), the route state is updated by a hysteretic threshold rule. Intuitively, a larger gap favors more conservative distillation, while a smaller gap allows stronger feature-level transfer. Let \(r_{e-1}\in\mathcal{R}\) denote the previous route state. We define the threshold set
\begin{equation}
\eta=\bigl(\eta_{nf}^{in},\eta_{nf}^{out},\eta_d^{in},\eta_d^{out},\eta_f^{out},\eta_f^{in}\bigr),
\end{equation}
with
\begin{equation}
\eta_{nf}^{in}\le \eta_{nf}^{out}\le \eta_d^{in}\le \eta_d^{out}\le \eta_f^{out}\le \eta_f^{in}.
\end{equation}
Let \(h_{e-1}\) be the current hold length and \(H\) the minimum hold length. If \(h_{e-1}<H\), the route state remains unchanged, with \(r_e=r_{e-1}\), to avoid excessively frequent route switching. Otherwise, the route is updated only when \(g_e\) crosses the corresponding entry threshold of a neighboring state; if \(g_e\) falls inside a hysteresis interval between the in/out thresholds, we retain the previous state, i.e., \(r_e=r_{e-1}\). The threshold-based mapping is
\begin{equation}
r_e=\Phi(g_e;\eta)=
\begin{cases}
3, & g_e\le \eta_{nf}^{in},\\
2, & \eta_{nf}^{out}\le g_e<\eta_d^{in},\\
1, & \eta_d^{out}\le g_e<\eta_f^{out},\\
0, & g_e\ge \eta_f^{in}.
\end{cases}
\end{equation}

\paragraph{Progressive weighting.}
Given the selected route \(r_e\), the logit-level and feature-level distillation weights are defined as
\begin{equation}
\begin{aligned}
w_l(e) &= \sigma\!\left(\frac{e-\beta_l(r_e)}{\tau_l}\right),\\
w_f(e) &= \sigma\!\left(\frac{e-\beta_f(r_e)}{\tau_f}\right)\,
\alpha_{\mathrm{gap}}(g_e)\,\mathbb{I}[r_e\neq 0].
\end{aligned}
\end{equation}
where \(\sigma(\cdot)\) denotes the sigmoid function, \(\beta_l(r_e)\) and \(\beta_f(r_e)\) are route-dependent onset hyperparameters controlling when logit-level and feature-level distillation begin to take effect, respectively, and \(\tau_l,\tau_f\) control the smoothness of the corresponding transitions. Detailed route-specific settings and implementation choices for these hyperparameters are given in the Appendix. The feature-level weight is further modulated by a gap-aware factor
\begin{equation}
\alpha_{\mathrm{gap}}(g_e)=
\begin{cases}
1, & g_e\le g_{\mathrm{low}},\\[2pt]
\dfrac{g_{\mathrm{high}}-g_e}{g_{\mathrm{high}}-g_{\mathrm{low}}}, & g_{\mathrm{low}}<g_e<g_{\mathrm{high}},\\[6pt]
0, & g_e\ge g_{\mathrm{high}},
\end{cases}
\end{equation}
where \(g_{\mathrm{low}}\) and \(g_{\mathrm{high}}\) are the lower and upper gap thresholds, respectively, controlling the progressive activation of feature-level distillation from suppressed to fully enabled as the teacher--student gap decreases and the two representations become more compatible over the course of training in a stable manner.

\paragraph{Distillation objective.}
Under the selected route configuration, the feature-level distillation objective is defined as
\begin{equation}
\tilde f_q^S = g_\phi(f_q^S), \qquad
\mathcal{L}_{\mathrm{feat\mbox{-}kd}}
=
\frac{1}{Q}\sum_{q=1}^{Q}
\left\|\tilde f_q^S - f_q^T\right\|_2^2,
\end{equation}
where \(g_\phi\) is a learnable projection head, \(f_q^S\) is the student embedding, \(f_q^T\) is the teacher feature, and \(\tilde f_q^S\) is the projected student feature in the teacher space. Logit-level knowledge distillation is defined by
\begin{equation}
\mathcal{L}_{\mathrm{logit\mbox{-}kd}}
=
\tau_{\mathrm{kd}}^2\, 
\mathrm{KL}
\left(
\mathrm{softmax}\left(\frac{\mathbf{E}^T}{\tau_{\mathrm{kd}}}\right)
\;\middle\|\;
\mathrm{softmax}\left(\frac{\mathbf{E}^S}{\tau_{\mathrm{kd}}}\right)
\right),
\end{equation}
where \(\mathbf{E}^T=[E_1^T,\ldots,E_{10}^T]\) and \(\mathbf{E}^S=[E_1^S,\ldots,E_{10}^S]\) denote the teacher and student digit-level evidence vectors, respectively, in which \(E_d^T\) and \(E_d^S\) represent the aggregated evidence scores for digit \(d\), obtained by summing question-level evidence across all questions querying that digit. The route-dependent distillation objective is
\begin{equation}
\mathcal{L}_{\mathrm{kd}}(e)
=
\lambda_f\,w_f(e)\,\mathcal{L}_{\mathrm{feat\mbox{-}kd}}
+
\lambda_l\,w_l(e)\,\mathcal{L}_{\mathrm{logit\mbox{-}kd}},
\end{equation}
where \(\lambda_f\) and \(\lambda_l\) are static trade-off coefficients balancing them.

\paragraph{Hierarchical Task Supervision.}
The student is further supervised by ground-truth labels at both the question and digit levels:
\begin{equation}
\begin{aligned}
\mathcal{L}_{\mathrm{que\mbox{-}ce}}
&=
-\frac{1}{Q}\sum_{q=1}^{Q}
\Big[y_q\log p_q +(1-y_q)\log(1-p_q)\Big],\\
\mathcal{L}_{\mathrm{digit\mbox{-}ce}}
&=
-\sum_{d=1}^{10}\mathbb{I}[d=d^\star]\,
\log\frac{\exp(E_d^S)}{\sum_{j=1}^{10}\exp(E_j^S)},
\end{aligned}
\end{equation}
where \(y_q\in\{0,1\}\) is the question-level deception label, \(d^\star\) is the ground-truth concealed digit, and \(\mathbb{I}[\cdot]\) is the indicator function.

\subsection{Overall objective and optimization}
\label{subsubsec:optimization}

The overall training objective of GPD is defined as
\begin{equation}
\mathcal{L}
=
 \mathcal{L}_{\mathrm{que\mbox{-}ce}}
+
\lambda_d \mathcal{L}_{\mathrm{digit\mbox{-}ce}}
+
\lambda_c \mathcal{L}_{\mathrm{kd}}(e),
\end{equation}
where \(\lambda_d\) and \(\lambda_c\) are trade-off coefficients. During training, the teacher is frozen, while the student network and the projection head \(g_\phi\) are optimized by backpropagation.

\begin{table*}[htbp]
\centering
\caption{Main comparison with state-of-the-art CMKD methods on MuDD. Results are reported as mean $\pm$ std over five runs with different random seeds. The best results are shown in bold, and the second-best results are underlined.}
\label{tab:main_results}
\setlength{\tabcolsep}{4.5pt}
\renewcommand{\arraystretch}{1.15}
\resizebox{\textwidth}{!}{
\begin{tabular}{lcccccccccc}
\toprule
\rowcolor{gray!12}
\multirow{2}{*}{Method} & \multirow{2}{*}{Params (M)} & \multirow{2}{*}{FLOPs/sample (M)}
& \multicolumn{4}{c}{Video modality}
& \multicolumn{4}{c}{Audio modality} \\
\cmidrule(lr){4-7} \cmidrule(lr){8-11}
\rowcolor{gray!8}
& &
& Top-1 & Top-2  & F1 & AUC
& Top-1 & Top-2  & F1 & AUC \\
\midrule
GENLIE
& \textbf{0.908} & 36.75
& \underline{26.8$\pm$5.0} & 35.3$\pm$5.7 & \underline{16.2$\pm$2.8} & \underline{54.4$\pm$1.4}
& 29.9$\pm$6.7 & 39.8$\pm$5.1 & 18.1$\pm$4.8 & 56.1$\pm$4.2 \\

XKD
& 1.237 & 49.32
& 23.6$\pm$6.5 & \underline{38.4$\pm$4.5} & 13.9$\pm$4.9 & 53.3$\pm$1.9
& 28.9$\pm$9.7 & 39.1$\pm$8.6 & 15.3$\pm$4.2 & 55.1$\pm$6.4 \\

SemBridge
& 1.456 & \textbf{34.16}
& 23.7$\pm$5.7 & 32.9$\pm$5.1 & 12.9$\pm$7.1 & 52.5$\pm$3.6
& 28.4$\pm$6.7 & \underline{39.9$\pm$6.3} & \underline{18.6$\pm$5.2} & \underline{58.3$\pm$2.9} \\

HKD-Emotion
& 1.075 & \textbf{34.16}
& 24.4$\pm$7.3 & 33.7$\pm$5.1 & 8.1$\pm$6.9 & 52.3$\pm$1.8
& 29.9$\pm$9.0 & 36.1$\pm$11.5 & 7.6$\pm$7.8 & 58.0$\pm$2.0 \\

DecomKD
& 1.535 & 61.78
& 24.5$\pm$6.0 & 32.1$\pm$8.8 & 5.8$\pm$5.7 & 52.8$\pm$2.5
& \underline{30.0$\pm$7.2} & 39.3$\pm$8.1 & 14.5$\pm$7.6 & 58.5$\pm$3.8 \\

CorrKD
& 1.075 & 42.40
& 22.3$\pm$2.3 & 33.7$\pm$4.7 & 11.0$\pm$5.1 & 52.6$\pm$2.0
& 27.6$\pm$5.6 & 38.4$\pm$6.7 & 18.3$\pm$3.3 & \underline{58.8$\pm$2.9} \\

CDGKD
& 1.209 & 48.22
& 24.5$\pm$5.8 & 37.6$\pm$6.8 & 13.7$\pm$5.9 & 53.5$\pm$2.0
& 27.6$\pm$6.6 & 39.1$\pm$7.7 & 15.1$\pm$8.2 & 56.7$\pm$2.4 \\

C2KD
& 0.944 & 37.69
& 23.8$\pm$2.3 & 33.1$\pm$5.6 & 12.7$\pm$3.2 & 52.4$\pm$3.1
& \underline{30.0$\pm$4.3} & 36.9$\pm$6.7 & 18.4$\pm$3.5 & 57.7$\pm$3.2 \\

MST-Distill
& 1.049 & \textbf{34.16}
& 23.1$\pm$2.3 & 34.6$\pm$3.2 & 8.5$\pm$4.9 & 51.1$\pm$3.1
& 28.5$\pm$4.4 & 37.6$\pm$7.6 & 17.5$\pm$4.9 & 57.6$\pm$4.2 \\

LogitStd
& \underline{0.921} & \underline{36.32}
& 24.6$\pm$4.4 & 30.8$\pm$6.3 & 5.5$\pm$6.0 & 52.6$\pm$3.1
& 28.4$\pm$6.7 & \underline{39.9$\pm$6.3} & 18.5$\pm$5.2 & 57.3$\pm$2.9 \\

\midrule
\rowcolor{gray!10}
\textbf{GPD (Ours)}
& \textbf{0.908} & 36.79
& \textbf{32.3$\pm$4.4} & \textbf{40.1$\pm$4.5} & \textbf{16.7$\pm$3.8} & \textbf{54.6$\pm$2.3}
& \textbf{35.3$\pm$4.9} & \textbf{43.8$\pm$6.6} & \textbf{20.1$\pm$2.3} & \textbf{59.9$\pm$1.85} \\
\bottomrule
\end{tabular}
}
\end{table*}
\section{Experiments}
% This section presents the experimental evaluation of GPD on MuDD. We first describe the experimental setting, including implementation details, data preprocessing, and evaluation protocol, and then compare GPD with representative CMKD baselines to assess its overall effectiveness. Next, we conduct a series of ablation studies to examine the roles of the two distillation objectives, progressive weighting, dynamic routing, and their interaction. Finally, we provide visualization-based analyses to further understand how GSR-guided distillation reshapes the student’s decision behavior.
\subsection{Experimental Setting}
\paragraph{Implementation details.}
All experiments are conducted on a single NVIDIA GeForce RTX 4090 GPU. We train GPD for 120 epochs using Adam with a batch size of 16, an initial learning rate of \(1\times10^{-4}\), and a weight decay of \(1\times10^{-4}\). Unless otherwise specified, experiments are conducted with random seed 42. For the main comparison with state-of-the-art CMKD methods, all methods are repeated over five runs with different random seeds, and the results are reported as mean $\pm$ std. For the student model, the hidden dimension and re-embedding dimension are set to 512 and 256, respectively, and the dropout rate is set to 0.3. The trade-off coefficients are set to \(\lambda_d=0.3\), \(\lambda_c=1.0\), \(\lambda_l=0.7\), and \(\lambda_f=0.2\), and the distillation temperature is set to \(\tau_{\mathrm{kd}}=2.0\). For gap-aware dynamic routing and progressive weighting, we set \(\mu=0.8\) and \(H=3\). The gap-aware weighting thresholds \((g_{\mathrm{low}}, g_{\mathrm{high}})\) are set to \((0.40,0.62)\) for video and \((0.46,0.76)\) for audio. The routing thresholds \((\eta_{nf}^{in},\eta_{nf}^{out},\eta_d^{in},\eta_d^{out},\eta_f^{out},\eta_f^{in})\) are set to (0.34, 0.40, 0.40, 0.48, 0.52, 0.60) for video and \((0.40, 0.46, 0.54, 0.60, 0.68, 0.76)\) for audio.

\paragraph{Preprocessing and feature extraction.}
For GSR, we apply a 2 Hz low-pass filter, downsample the signal from 256 Hz to 32 Hz, and resize each segment to 448 sampling points by truncation or zero-padding. The processed GSR is then fed into a teacher encoder adapted from~\cite{ref47}, producing an 80-dimensional representation. Detailed architectural modifications are provided in the Appendix. For the visual modality, following GENLIE~\cite{ref51}, we detect and align face regions based on facial landmarks, divide the aligned video into temporal segments, sample 16 frames per segment, and extract 768-dimensional visual features using VideoMAEv2. For the audio modality, frame-level acoustic features are extracted using pretrained WavLM~\cite{ref52}, and local average pooling is applied according to the visual segment boundaries to obtain temporally aligned 768-dimensional audio features.

\paragraph{Evaluation protocol.}
We evaluate all methods under 5-fold cross-validation, with detailed split information provided in the Appendix. For question-level binary deception detection, we report F1 score and AUC, treating deception as the positive class. F1 measures the balance between precision and recall, while AUC measures class separability across decision thresholds. For trial-level concealed-digit identification, we report Top-1 and Top-2 accuracy, indicating whether the ground-truth concealed digit is ranked within the top one or top two predictions, respectively. Since the numbers of samples differ slightly across folds, final results are reported as sample-weighted averages over the five folds.
\subsection{Main Results}
\paragraph{Baseline Models.}
We compare GPD against nine representative state-of-the-art CMKD methods and the original non-distilled GENLIE baseline. To isolate the effect of different distillation strategies, all methods are implemented with the same student backbone and evaluated under the same 5-fold cross-validation protocol. GENLIE~\cite{ref51} serves as the non-distilled student backbone. The CMKD baselines include XKD~\cite{ref21}, which performs self-supervised audio--visual distillation; SemBridge~\cite{ref48}, which uses contrastive cross-modal distillation; HKD-Emotion~\cite{ref49}, a hierarchical distillation framework; DecomKD~\cite{ref17}, a decomposed distillation framework; CorrKD~\cite{ref50}, a correlation-decoupled distillation framework; CDGKD~\cite{ref41}, a densely guided distillation framework; C2KD~\cite{ref40}, which adopts bidirectional cross-modal distillation; MST-Distill~\cite{ref42}, a multi-stage teacher distillation framework; and LogitStd~\cite{ref20}, which improves distillation via logit standardization. The comparison results are reported in Table~\ref{tab:main_results}.

As shown in Table~\ref{tab:main_results}, GPD achieves the best overall performance among all compared methods. In particular, its gains are more pronounced on digit-level prediction, where it achieves the highest Top-1 and Top-2 accuracy. This suggests that the proposed digit-level evidence transfer is effective in conveying the teacher’s concealed-digit discrimination ability to the non-contact student. At the same time, GPD also maintains the strongest F1/AUC results for question-level deception detection, indicating that the transferred knowledge remains beneficial at the finer-grained binary decision level. Another notable observation is that several existing CMKD baselines fail to consistently outperform the plain GENLIE model, and some even exhibit negative transfer. This suggests that under the large gap between physiological GSR signals and non-contact modalities, existing distillation strategies may be less effective because they typically rely on predefined transfer designs and do not explicitly model the evolving teacher--student discrepancy during training. In contrast, GPD models both what knowledge is transferred and how the transfer process is scheduled during training. The feature-level and logit-level objectives provide complementary supervision on intermediate representations and prediction-level signals, while dynamic routing and progressive weighting allow the transfer path and supervision strength to adapt to the evolving teacher--student discrepancy. For completeness, the standalone performance of the GSR teacher is reported in the Appendix. It is also worth noting that the gains of GPD are not obtained by increasing model size.
\begin{table}[t]
\centering
\caption{Ablation study of different components in GPD on MuDD. Prog. wt. denotes progressive weighting.}
\label{tab:ablation_results}
\setlength{\tabcolsep}{3pt}
\renewcommand{\arraystretch}{1.15}
\scriptsize
\resizebox{\columnwidth}{!}{
\begin{tabular}{lcccccccc}
\toprule
\rowcolor{gray!12}
\multirow{2}{*}{Method}
& \multicolumn{4}{c}{Video modality}
& \multicolumn{4}{c}{Audio modality} \\
\cmidrule(lr){2-5} \cmidrule(lr){6-9}
\rowcolor{gray!8}
& Top-1 & Top-2 & F1 & AUC
& Top-1 & Top-2 & F1 & AUC \\
\midrule
GENLIE
& 26.92 & 35.38 & 16.30 & 53.14
& 29.86 & 40.77 & 18.94 & 59.03 \\

w/o \(\mathcal{L}_{\mathrm{logit\mbox{-}kd}}\)
& 26.92 & 37.69 & 17.00 & 52.13
& 30.77 & 41.27 & 18.50 & 57.48 \\

w/o \(\mathcal{L}_{\mathrm{feat\mbox{-}kd}}\)
& 25.38 & 36.92 & 16.70 & 51.46
& 32.31 & 43.08 & 19.70 & 57.66 \\

w/o prog. wt.
& 23.08 & 35.38 & 16.72 & 51.89
& 30.77 & 42.47 & 20.05 & 59.25 \\

\rowcolor{gray!10}
GPD
& \textbf{32.31} & \textbf{40.00} & \textbf{17.43} & \textbf{53.55}
& \textbf{35.38} & \textbf{43.84} & \textbf{20.63} & \textbf{59.30} \\
\bottomrule
\end{tabular}
}
\end{table}
\subsection{Component and Strategy Ablation}
\paragraph{Q1: Are the two distillation objectives and progressive weighting necessary?}
We first examine the contribution of each core component in GPD. Starting from the original student baseline GENLIE, we compare the full model against three variants: \textit{w/o} \(\mathcal{L}_{\mathrm{logit\mbox{-}kd}}\), \textit{w/o} \(\mathcal{L}_{\mathrm{feat\mbox{-}kd}}\), and \textit{w/o progressive weighting}. As shown in Table~\ref{tab:ablation_results}, GPD achieves the best overall performance in both the video and audio modalities. Removing \(\mathcal{L}_{\mathrm{logit\mbox{-}kd}}\) or \(\mathcal{L}_{\mathrm{feat\mbox{-}kd}}\) consistently degrades the results, indicating that the two distillation objectives provide complementary supervision. Removing progressive weighting also leads to inferior performance, confirming its importance for stable knowledge transfer across the large teacher--student modality gap.

\paragraph{Q2: Which progressive weighting strategy is most effective?}
We compare four progressive weighting strategies while keeping the architecture and the two distillation objectives unchanged: sigmoid, which increases the weights smoothly with a slow start and saturation; linear, which increases them at a constant rate; step, which activates them abruptly at predefined stages; and cosine, which follows a smooth nonlinear schedule. As shown in Table~\ref{tab:progressive_weighting}, the sigmoid schedule achieves the best overall performance. We attribute this to its smooth transition, which is better suited to coordinating the two complementary distillation objectives. Compared with linear, step, and cosine schedules, sigmoid weighting provides a more stable balance between \(\mathcal{L}_{\mathrm{feat\mbox{-}kd}}\) and \(\mathcal{L}_{\mathrm{logit\mbox{-}kd}}\) throughout training.

\begin{table}[h]
\centering
\caption{Comparison of different progressive weighting schedules on MuDD.}
\label{tab:progressive_weighting}
\setlength{\tabcolsep}{3pt}
\renewcommand{\arraystretch}{1.15}
\scriptsize
\resizebox{\columnwidth}{!}{
\begin{tabular}{lcccccccc}
\toprule
\rowcolor{gray!12}
\multirow{2}{*}{Schedule}
& \multicolumn{4}{c}{Video modality}
& \multicolumn{4}{c}{Audio modality} \\
\cmidrule(lr){2-5} \cmidrule(lr){6-9}
\rowcolor{gray!8}
& Top-1 & Top-2 & F1 & AUC
& Top-1 & Top-2 & F1 & AUC \\
\midrule
Linear 
& 30.77 & \textbf{41.54} & 17.27 & 52.67
& 33.08 & 38.46 & 18.97 & 57.07 \\

Step   
& 30.00 & \textbf{41.54} & \textbf{17.74} & 52.80
& 34.62 & 40.76 & 19.22 & 57.41 \\

Cosine 
& 30.00 & 39.23 & 17.53 & 52.70
& 33.08 & 40.00 & 19.06 & 58.20 \\

\rowcolor{gray!10}
Sigmoid  
& \textbf{32.31} & 40.00 & 17.43 & \textbf{53.55}
& \textbf{35.38} & \textbf{43.84} & \textbf{20.63} & \textbf{59.30} \\
\bottomrule
\end{tabular}
}
\end{table}

\paragraph{Q3: Is dynamic routing better than fixed route configurations?}
We compare the full model with dynamic routing (Dyn.) against four fixed route configurations: no-feature (No-feat.), logit-first, joint, and feature-first (Feat.-first). Table~\ref{tab:path_ablation} shows that dynamic routing achieves the best overall performance. We attribute this to the fact that the teacher--student gap evolves during training, making a single fixed route suboptimal across stages. When the gap is large, more conservative routes are preferable, whereas smaller gaps allow stronger feature-level transfer. Dynamic routing adapts the transfer path to this changing optimization state, while fixed configurations impose the same distillation order throughout training. This interpretation is supported by the behavior of individual fixed routes: the \textit{joint} route may introduce conflicting supervision when the teacher--student gap is still large, while \textit{feature-first} can be too aggressive and \textit{logit-first} too conservative for effective transfer throughout training. Although \textit{no-feature} remains competitive on some video metrics, its lack of feature-level guidance limits overall performance, especially for audio.

\begin{table}[t]
\centering
\caption{Comparison of different distillation routes on MuDD.}
\label{tab:path_ablation}
\setlength{\tabcolsep}{3pt}
\renewcommand{\arraystretch}{1.15}
\scriptsize
\resizebox{\columnwidth}{!}{
\begin{tabular}{lcccccccc}
\toprule
\rowcolor{gray!12}
\multirow{2}{*}{Route}
& \multicolumn{4}{c}{Video modality}
& \multicolumn{4}{c}{Audio modality} \\
\cmidrule(lr){2-5} \cmidrule(lr){6-9}
\rowcolor{gray!8}
& Top-1 & Top-2 & F1 & AUC
& Top-1 & Top-2 & F1 & AUC \\
\midrule

No-feat. 
& 30.77 & \textbf{41.54} & \textbf{17.68} & 52.69
& 34.62 & 41.54 & 19.75 & 57.47 \\

Feat.-first
& 31.31 & 39.23 & 17.31 & 52.81
& 33.08 & 42.31 & 19.42 & 58.14 \\

Logit-first
& 31.54 & 40.77 & 17.47 & 52.46
& 33.08 & 39.23 & 19.14 & 56.73 \\

Joint  
& 30.77 & 39.31 & 17.46 & 52.66
& 32.31 & 42.31 & 17.92 & 58.13 \\

\rowcolor{gray!10}
Dyn.   
& \textbf{32.31} & 40.00 & 17.43 & \textbf{53.55}
& \textbf{35.38} & \textbf{43.84} & \textbf{20.63} & \textbf{59.30}\\

\bottomrule
\end{tabular}
}
\end{table}

\subsection{Interaction Between Routing and Weighting}
\paragraph{Q4: Are dynamic routing and progressive weighting complementary?}
To decouple the effects of routing and progressive weighting, we toggle the routing mechanism on and off under different weighting schedules. As shown in Table~\ref{tab:routing_weight_decouple}, dynamic routing works best when combined with sigmoid weighting. A possible reason is that the two strategies regulate different aspects of the transfer process: routing determines which route configuration is used at a given stage, whereas weighting controls how strongly the selected route influences optimization. Under linear, step, and cosine schedules, the benefit of routing becomes less consistent, suggesting weaker coordination between route selection and distillation strength. Overall, these results indicate that effective transfer depends not only on selecting an appropriate route configuration, but also on pairing it with a weighting strategy that supports stable optimization.

\begin{table}[t]
\centering
\caption{Decoupled ablation of routing and weighting on MuDD. ``Wt.'' and ``Rt.'' denote weighting schedule and route configuration, while ``On'' and ``Off'' indicate whether dynamic routing is enabled. For each weighting schedule, the better result between On and Off is highlighted in bold.}
\label{tab:routing_weight_decouple}
\setlength{\tabcolsep}{2.5pt}
\renewcommand{\arraystretch}{1.15}
\tiny
\resizebox{\columnwidth}{!}{
\begin{tabular}{llcccccccc}
\toprule
\rowcolor{gray!12}
\multirow{2}{*}{Wt.} & \multirow{2}{*}{Rt.}
& \multicolumn{4}{c}{Video modality}
& \multicolumn{4}{c}{Audio modality} \\
\cmidrule(lr){3-6} \cmidrule(lr){7-10}
\rowcolor{gray!8}
& & Top-1 & Top-2 & F1 & AUC
& Top-1 & Top-2 & F1 & AUC \\
\midrule
Sig. & On  & \textbf{32.31} & 40.00 & \textbf{17.43} & \textbf{53.55}
     & \textbf{35.38} & \textbf{43.84} & \textbf{20.63} & \textbf{59.30} \\
     & Off & 30.77 & \textbf{42.31} & 17.37 & 52.66
     & 33.08 & 42.31 & 19.42 & 58.14 \\
\midrule
Lin. & On  & 30.77 & \textbf{41.54} & \textbf{17.27} & \textbf{52.67}
     & \textbf{33.08} & 38.46 & \textbf{18.97} & \textbf{57.07} \\
     & Off & \textbf{31.54} & 40.77 & 17.26 & 52.52
     & 31.54 & \textbf{41.54} & 17.77 & 56.68 \\
\midrule
Step & On  & 30.00 & \textbf{41.54} & 17.74 & \textbf{52.80}
     & \textbf{34.62} & \textbf{40.76} & 19.22 & 57.41 \\
     & Off & \textbf{30.77} & 40.77 & \textbf{17.94} & 52.77
     & 32.30 & 40.00 & \textbf{20.02} & \textbf{57.68} \\
\midrule
Cos. & On  & 30.00 & 39.23 & \textbf{17.53} & \textbf{52.70}
     & \textbf{33.08} & 40.00 & \textbf{19.06} & \textbf{58.20} \\
     & Off & \textbf{30.77} & \textbf{40.00} & 17.21 & 52.59
     & 32.31 & \textbf{40.77} & 18.71 & 58.02 \\
\bottomrule
\end{tabular}
}
\end{table}
\subsection{Sensitivity of Dynamic Routing}
\begin{figure}[h]
    \centering
    \begin{subfigure}[t]{0.48\columnwidth}
        \centering
        \includegraphics[width=\textwidth]{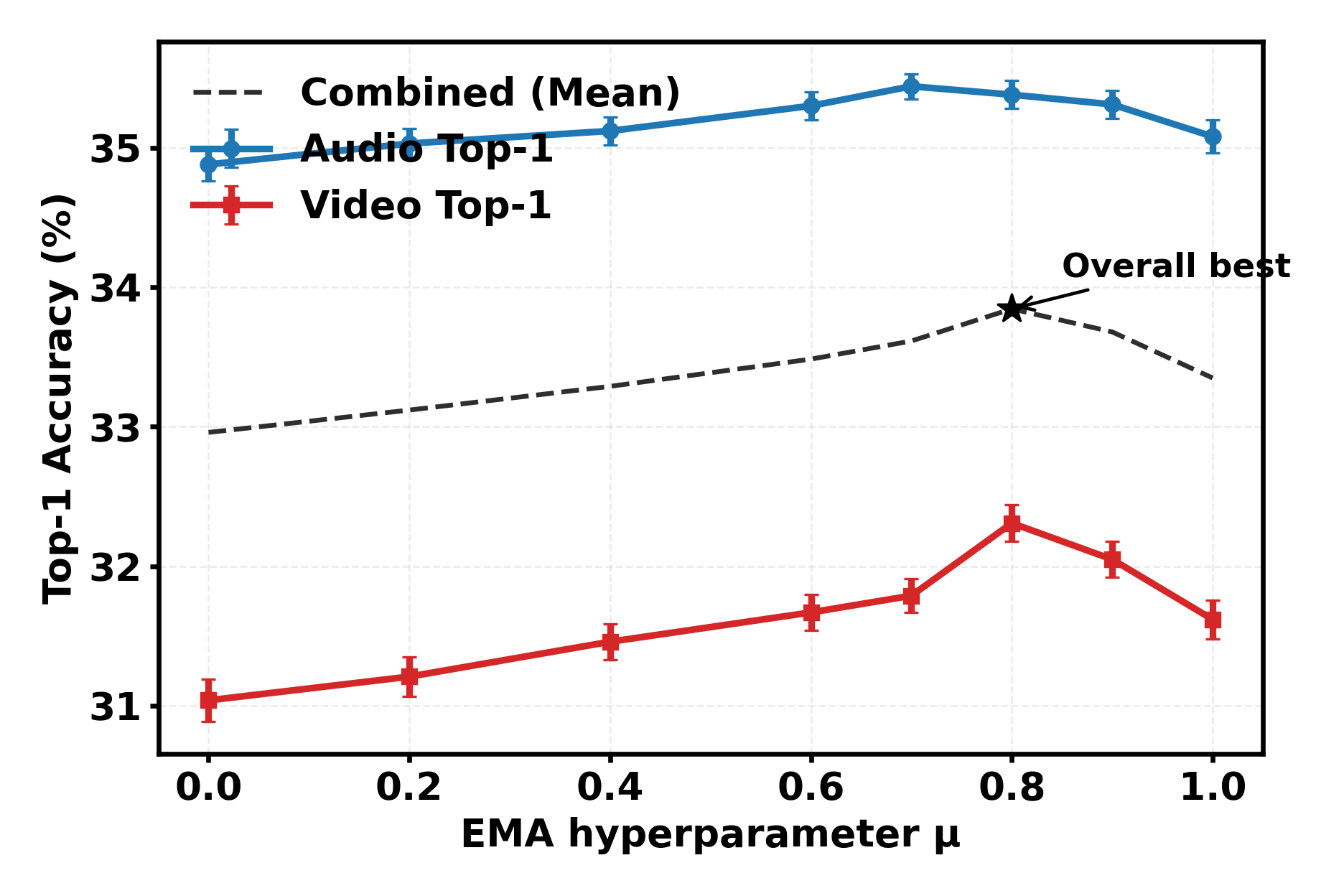}
        \caption{EMA momentum \(\mu\).}
        \label{fig:ema_sensitivity}
    \end{subfigure}
    \hfill
    \begin{subfigure}[t]{0.48\columnwidth}
        \centering
        \includegraphics[width=\textwidth]{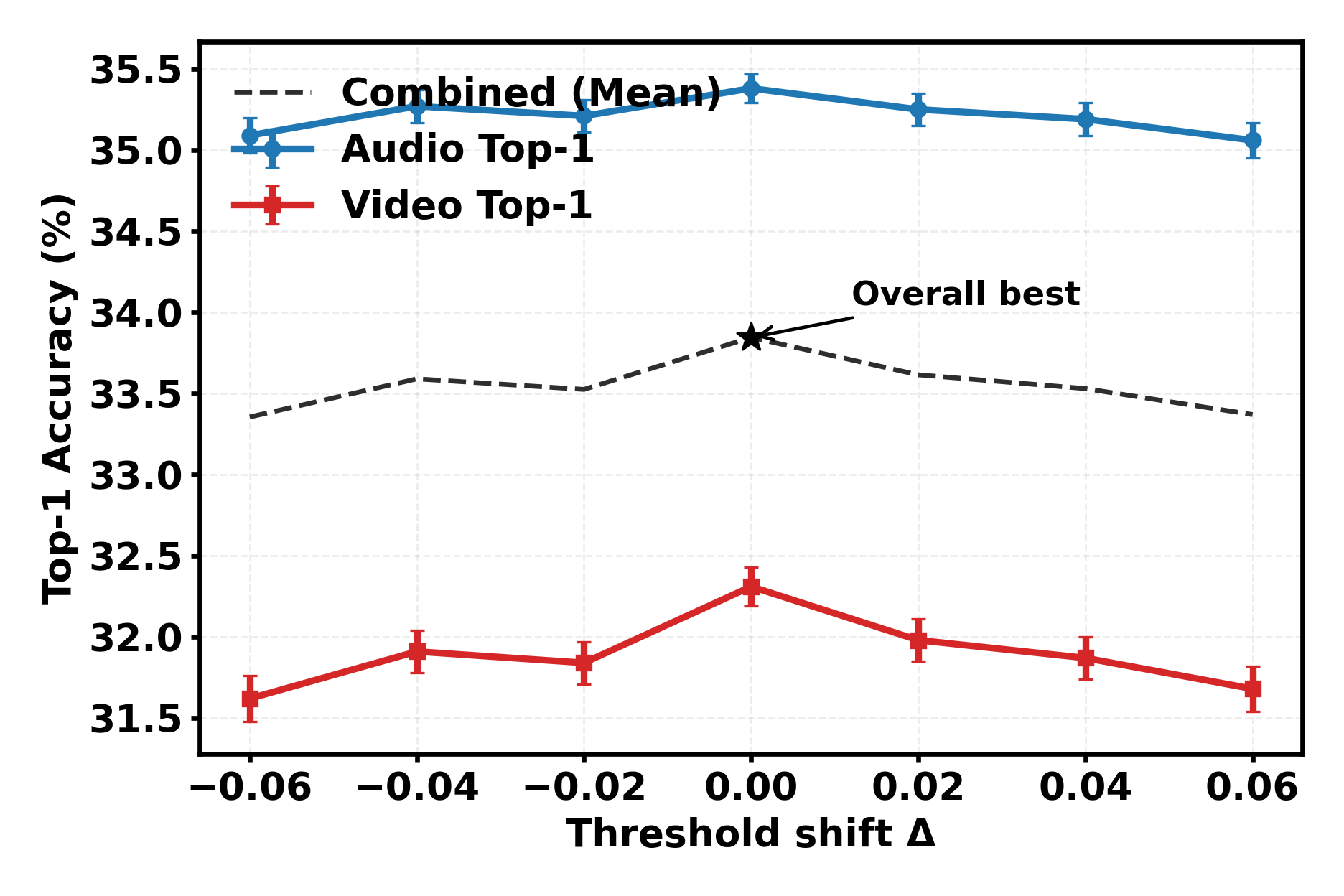}
        \caption{Threshold shift \(\Delta\).}
        \label{fig:threshold_sensitivity}
    \end{subfigure}
    \caption{Sensitivity analysis of dynamic routing. Left: varying the EMA momentum \(\mu\). Right: uniformly shifting all routing thresholds by \(\Delta\).}
    \label{fig:routing_sensitivity}
\end{figure}

\paragraph{Q5: How sensitive is dynamic routing to key hyperparameters?}
We vary the EMA momentum \(\mu\) in Eq.~(4) to examine how the smoothness of the routing gate affects performance. As shown in Fig.~\ref{fig:ema_sensitivity}, both branches improve as \(\mu\) increases from \(0\), and the best overall mean performance is achieved at \(\mu=0.8\). The video branch peaks at \(\mu=0.8\), while the audio branch peaks slightly earlier around \(\mu=0.7\), followed by a mild decline. When \(\mu\) is too small, the gap estimate becomes sensitive to noisy epoch-level fluctuations and may trigger unstable route switching; when \(\mu\) is too large, the gate becomes overly smooth and less responsive. This suggests that a moderately large EMA momentum provides a good balance between stability and adaptability. To examine whether the routing policy depends on overly precise threshold tuning, we uniformly shift all routing thresholds of each modality by a shared offset \(\Delta\). As shown in Fig.~\ref{fig:threshold_sensitivity}, the best overall mean performance is achieved at \(\Delta=0\), while small perturbations around this point cause only minor degradation. As the thresholds deviate further from the original setting, the performance generally declines, indicating that excessive threshold shifts weaken effective transfer. These results suggest that the routing mechanism is locally robust rather than brittle to threshold selection.

\subsection{Effect of Distillation on Student Behavior}
\begin{figure}[t]
    \centering
    \begin{subfigure}[t]{0.48\columnwidth}
        \centering
        \includegraphics[width=\textwidth]{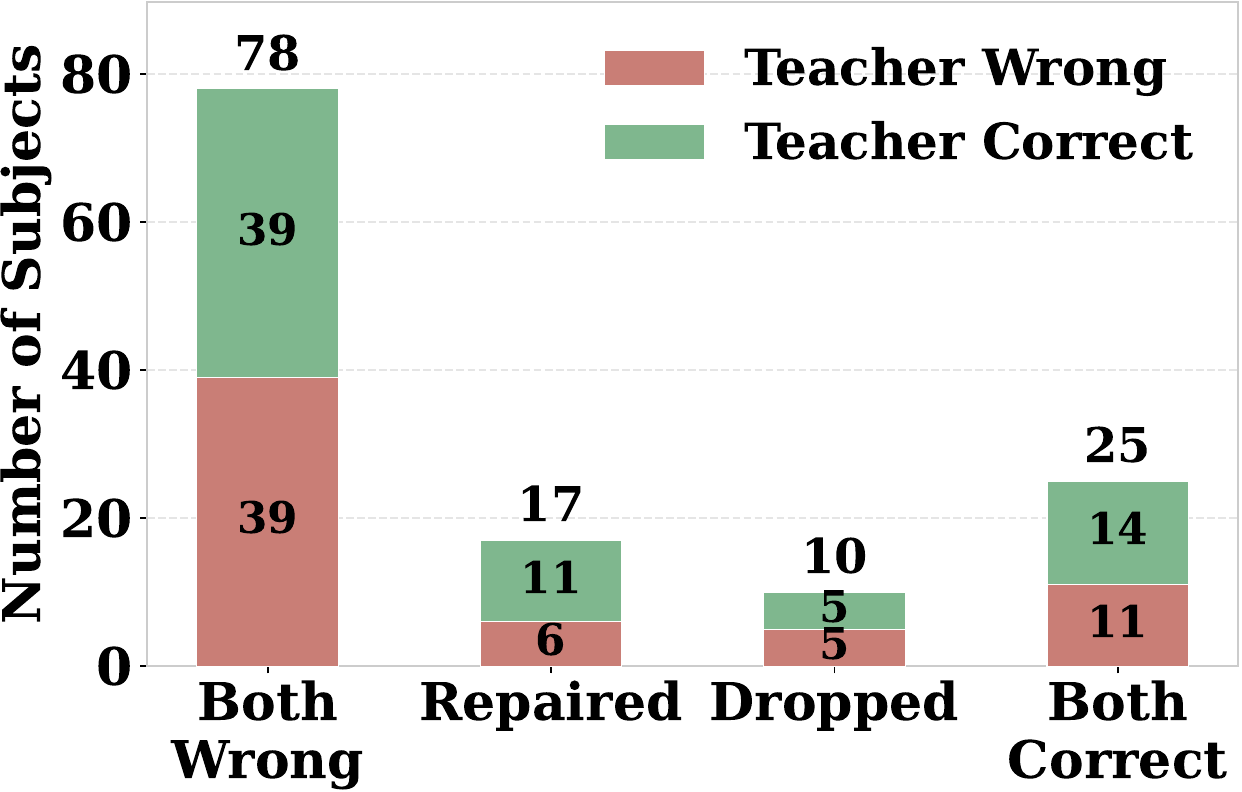}
        \caption{Video transition.}
        \label{fig:video_bar}
    \end{subfigure}
    \hfill
    \begin{subfigure}[t]{0.48\columnwidth}
        \centering
        \includegraphics[width=\textwidth]{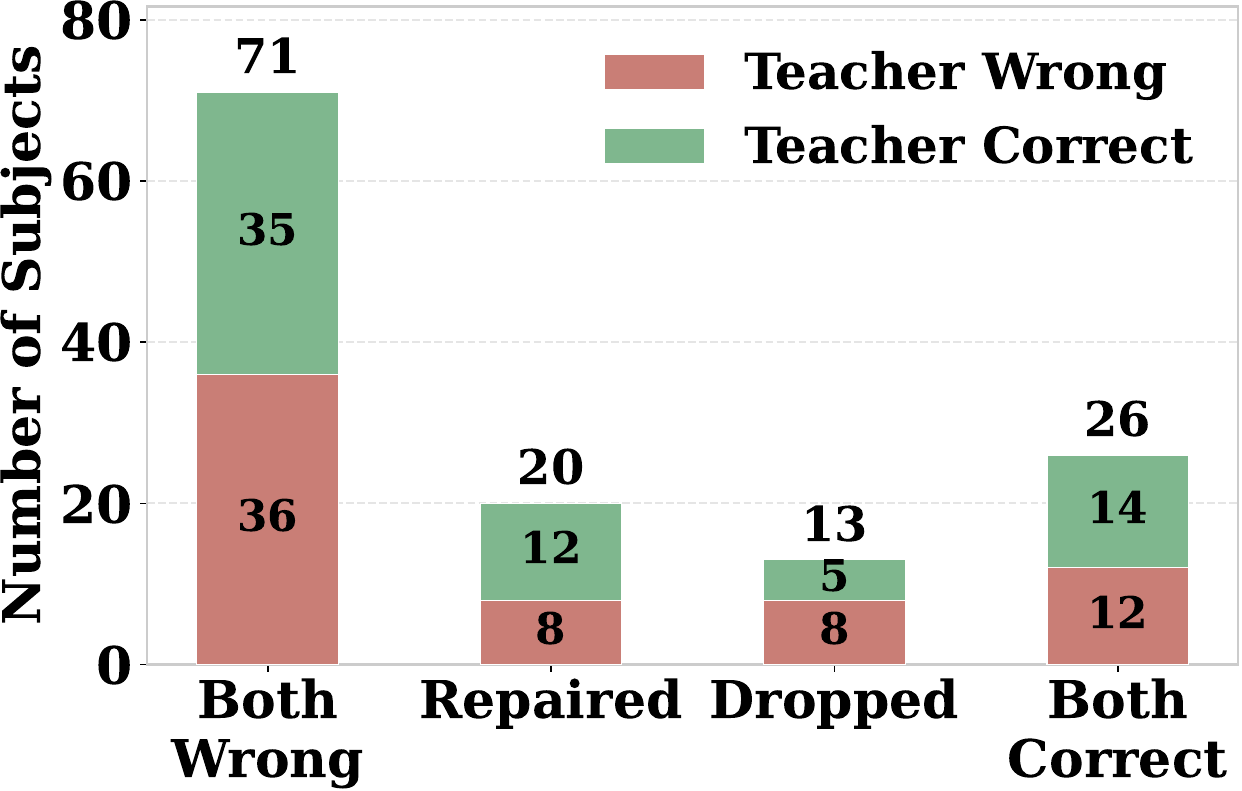}
        \caption{Audio transition.}
        \label{fig:audio_bar}
    \end{subfigure}

    \vspace{0.3em}

    \begin{subfigure}[t]{0.48\columnwidth}
        \centering
        \includegraphics[width=\textwidth]{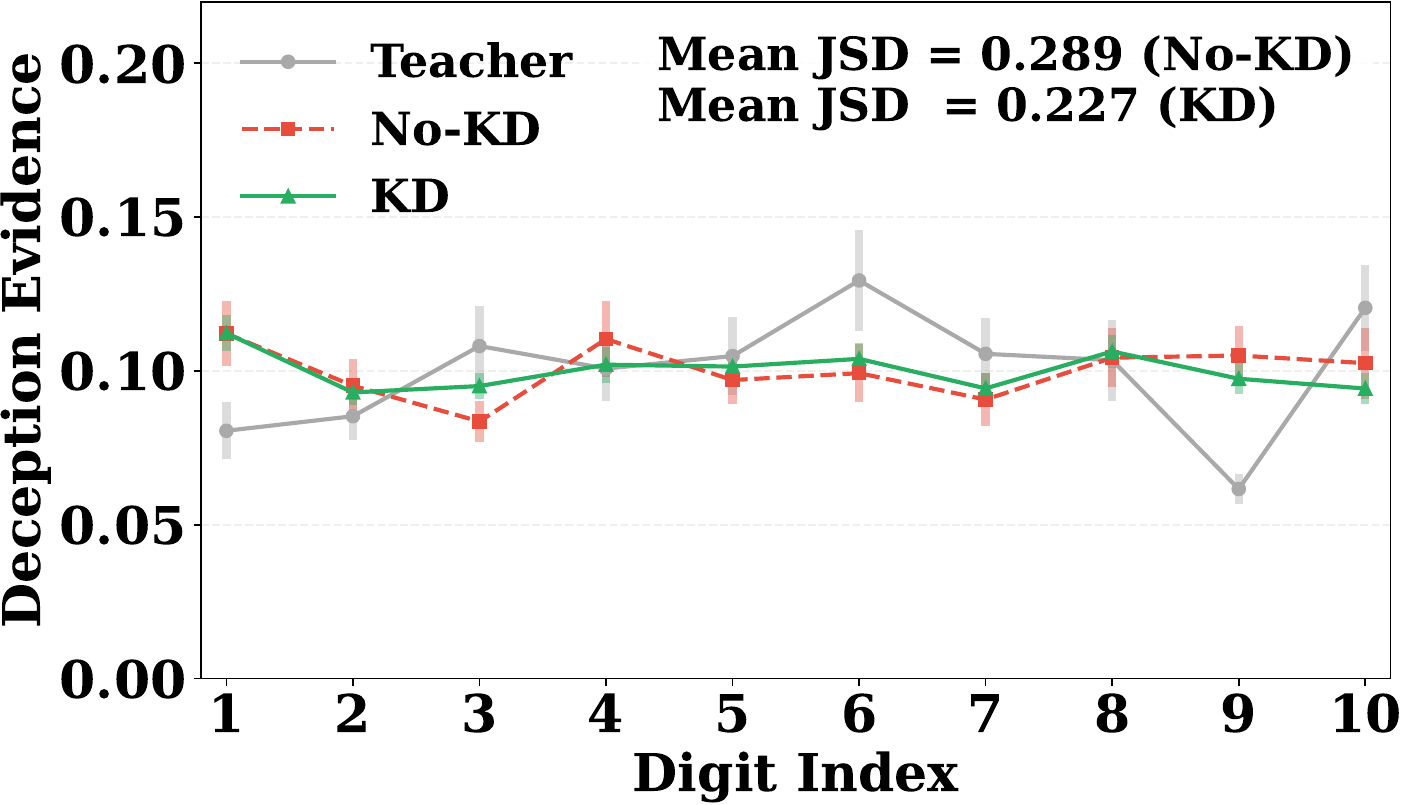}
        \caption{Video response curve.}
        \label{fig:video_jsd}
    \end{subfigure}
    \hfill
    \begin{subfigure}[t]{0.48\columnwidth}
        \centering
        \includegraphics[width=\textwidth]{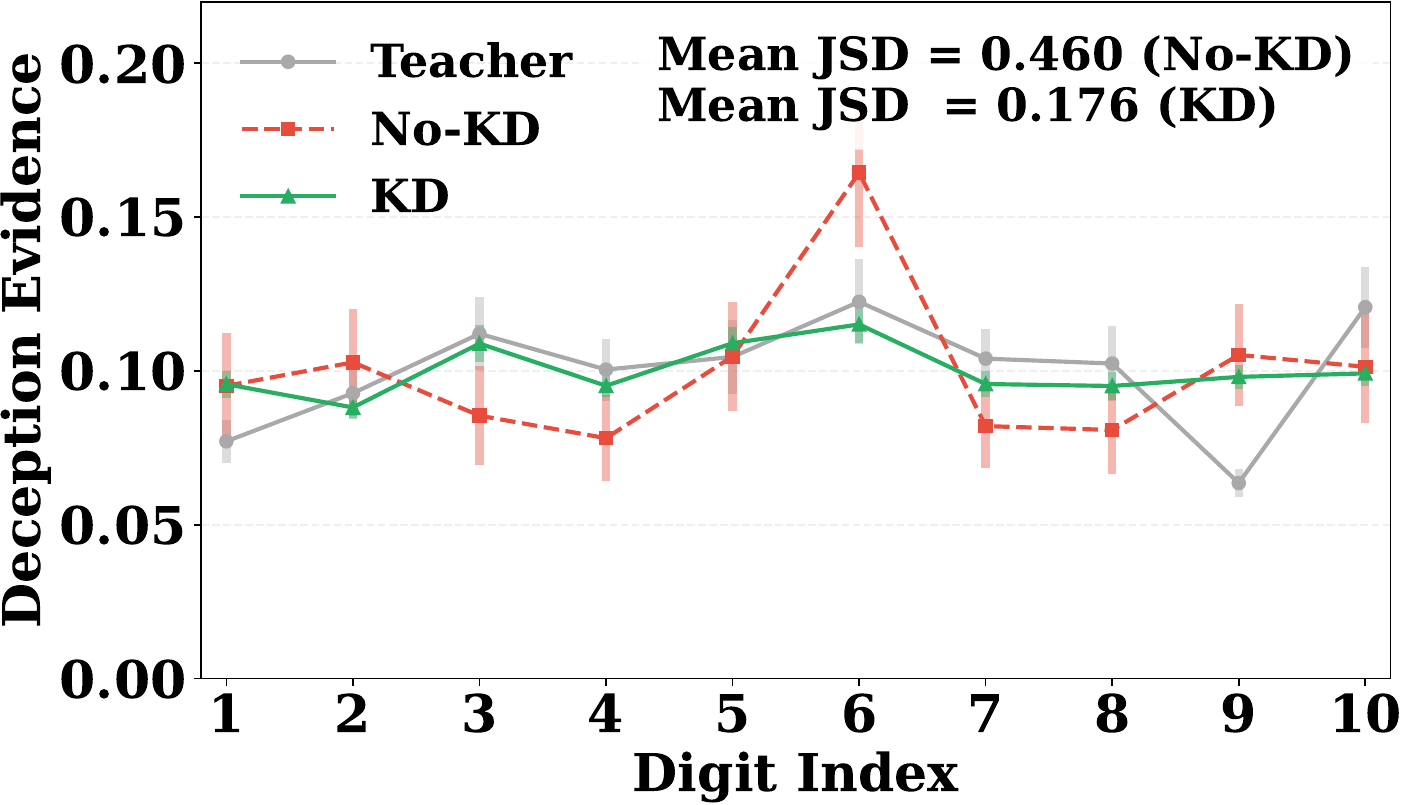}
        \caption{Audio response curve.}
        \label{fig:audio_jsd}
    \end{subfigure}

    \caption{Effect of distillation on student behavior in the video and audio modalities.}
    \label{fig:behavior_vis}
\end{figure}

\paragraph{Q6: How does distillation affect the student’s decision behavior?}
We analyze both video and audio students using prediction-transition statistics and representative response curves, as shown in Fig.~\ref{fig:behavior_vis}. The transition plots summarize prediction changes before and after distillation, including \textit{both-wrong}, \textit{repaired}, \textit{dropped}, and \textit{both-correct} cases, each further divided according to whether the teacher prediction is correct. The response curves show the distribution of digit-level evidence over the 10 candidate digits within each trial, reflecting how the student organizes trial-level evidence before and after distillation.
As shown in Fig.~\ref{fig:behavior_vis}, repaired samples outnumber dropped ones in both modalities, indicating that distillation improves the student more often than it harms it. Moreover, most repaired cases are teacher-consistent (\(11/17\) for video and \(12/20\) for audio), suggesting that the gains mainly come from effective teacher guidance rather than incidental prediction changes. Some dropped cases are also teacher-consistent, indicating that distillation can occasionally pull the student toward the teacher even when this does not improve the ground-truth decision. For audio, dropped cases contain a larger proportion of teacher-wrong samples (\(8/13\)), suggesting higher sensitivity to local teacher noise.
The response curves further show that the student after distillation aligns more closely with the teacher, especially around salient peaks and in the overall response trend. This suggests that distillation improves not only final predictions but also the student’s trial-level evidence organization, making its responses more consistent with the physiological teacher over trials.

\section{Conclusion}
We addressed non-contact deception detection from the perspective of physiological-signal-guided cross-modal distillation. Specifically, we introduced MuDD, a large-scale multimodal deception detection dataset collected under the Guilty Knowledge Test paradigm, and proposed GPD, a GSR-guided progressive distillation framework that uses gap-aware dynamic routing and progressive weighting to adapt both the transfer pathway and the transfer strength according to the evolving teacher--student gap.
Extensive experiments show that GPD effectively performs joint modeling of the two tasks and consistently outperforms representative cross-modal knowledge distillation (CMKD) baselines. These results indicate that GSR contains transferable and highly informative deception-related knowledge, and that modeling cross-modal distillation as a dynamic rather than predefined process enables more effective knowledge transfer under substantial modality discrepancy.

A limitation of the current work is that it does not explicitly model the alignment between physiological responses in GSR and overt deception cues in non-contact modalities. These cross-modal relations are currently captured implicitly through knowledge transfer. Future work will explore explicit alignment modeling to improve both interpretability and detection performance. 
% \section*{Data Availability}
% The MuDD dataset is not publicly available due to privacy and ethical restrictions associated with human-subject multimodal recordings, but may be made available to qualified researchers with institutional affiliations upon reasonable request to the corresponding author, subject to applicable ethical approval and data use requirements.

\bibliographystyle{ACM-Reference-Format}
\bibliography{sample-base}

\end{document}